\documentclass[10pt,conference]{IEEEtran}
\IEEEoverridecommandlockouts
\usepackage[T1]{fontenc}              
\usepackage{newtxtext,newtxmath}      
\usepackage{microtype}                

\usepackage{cite}

\usepackage{amsmath,amssymb,amsfonts}
\usepackage{algorithmic}
\usepackage[ruled]{algorithm2e}
\usepackage{tikz}
\usepackage{graphicx}
\usepackage{textcomp}
\usepackage{booktabs}
\usepackage{siunitx}
\usepackage{multirow}
\usepackage{xcolor}
\usepackage{caption}
\usepackage{bm}
\def\BibTeX{{\rm B\kern-.05em{\sc i\kern-.025em b}\kern-.08em
    T\kern-.1667em\lower.7ex\hbox{E}\kern-.125emX}}

\usepackage{booktabs}
\usepackage{siunitx}
\newcommand{\ti}[1]{\textit{#1}}  

\begin{document}

\title{Vehicle-to-Infrastructure Collaborative Spatial Perception via Multimodal Large Language Models}

\author{\IEEEauthorblockN{Kimia Ehsani and Walid Saad}
\IEEEauthorblockA{Bradley Department of Electrical and Computer Engineering, Virginia Tech, Alexandria, VA, USA \\
Emails: \{kimiaehsani,walids\}@vt.edu}
\thanks{This research was supported by the US National Science Foundation under Grant CNS-2225511.}
}

\maketitle

\begin{abstract}

Accurate prediction of communication link quality metrics is essential for  vehicle-to-infrastructure (V2I) systems, enabling smooth handovers, efficient beam management, and reliable low-latency communication. The increasing availability of sensor data from modern vehicles motivates the use of multimodal large language models (MLLMs) because of their adaptability across tasks and reasoning capabilities. However, MLLMs inherently lack three‑dimensional spatial understanding. To overcome this limitation, a lightweight, plug-and-play bird’s-eye view (BEV) injection connector is proposed.
In this framework, a BEV of the environment is constructed by collecting sensing data from neighboring vehicles. This BEV representation is then fused with the ego vehicle’s input  to provide spatial context for the large language model. To support realistic multimodal learning, a co-simulation environment combining CARLA simulator and MATLAB-based ray tracing is developed to generate RGB, LiDAR, GPS, and wireless signal data across varied scenarios. Instructions and ground-truth responses are programmatically extracted from the ray-tracing outputs. Extensive experiments are conducted across three V2I link prediction tasks: line-of-sight (LoS) versus non-line-of-sight (NLoS) classification, link availability, and blockage prediction. Simulation results show that the proposed BEV injection framework consistently improved performance across all tasks. The results indicate that, compared to an ego-only baseline, the proposed approach improves the macro-average of the accuracy metrics by up to 13.9\%. The results also show that this performance gain increases by up to 32.7\% under challenging rainy and nighttime conditions, confirming the robustness of the framework in adverse settings.

 \end{abstract}
 \vspace{0.5em}
\section{Introduction}

Multimodal sensing is a key enabler of data-intensive applications such as extended reality (XR), connected and autonomous vehicles (CAVs), and digital twins in 6G. \cite{saad_artificial_2025, chaccour_joint_2024} By leveraging complementary data streams, multimodal sensing can deliver the high-fidelity environmental understanding and robustness needed to meet 6G’s stringent requirements. Particularly, 6G vehicular networks can potentially harness rich, multimodal data collected directly from autonomous vehicles equipped by various types
of sensors that can enhance not only communication but also real-time environmental perception. However, leveraging multimodal sensing in vehicular networks faces a number of challenges that include heterogeneous sensor alignment and resource-efficient data fusion.

\subsection{Related Works}

Recent works have investigated the use of multimodal sensing to enhance wireless communication by integrating data from radar, LiDAR, and vision sensors.\cite{cui_sensing-assisted_2024, tian_multimodal_2023, charan_vision-aided_2021, ali_passive_2020} For example, the works in \cite{cui_sensing-assisted_2024} and \cite{tian_multimodal_2023} used LiDAR, radar,
RGB, and GPS data to improve beam prediction accuracy. In \cite{charan_vision-aided_2021}, the authors 
propose a vision-aided bimodal solution for  blockage prediction and user handoff. 
Similarly, in \cite{ali_passive_2020} the authors studied the use of passive radar to assist millimeter-wave (mmWave) beamforming by extracting spatial features from automotive radar signals.
However, the solutions of \cite{cui_sensing-assisted_2024, tian_multimodal_2023,charan_vision-aided_2021, ali_passive_2020} depend on task-specific fusion chains tailored to particular modal combinations.  As a result, adding another modality typically entails complete end-to-end retraining, and supporting diverse tasks demands architectural modifications. These issues significantly undermine the scalable deployment of 6G networks, which demand seamless integration and rapid adaptability to evolving modalities and use-case requirements. This lack of a unified end-to-end fusion framework for multimodal sensing has motivated the use of  large language models (LLMs), which provide a pre-trained universal backbone that can be quickly fine-tuned with lightweight modules across diverse communication downstream tasks. \cite{zheng_beamllm_2025, liu_llm4cp_2024, zhang_port-llm_2025}

LLMs can be effective in  few-shot generalization. As a result, a number of recent works applied LLMs to a variety of wireless communication tasks, including beam prediction \cite{zheng_beamllm_2025}, channel prediction\cite{liu_llm4cp_2024}, and port prediction \cite{zhang_port-llm_2025}. Originally developed for natural language processing (NLP), these models have since been extended into multimodal LLMs, supporting seamless integration of heterogeneous inputs. \cite{liu2023visualinstructiontuning}
For instance, The authors in \cite{zheng_beamllm_2025} developed a vision language model for beam prediction, while he work in \cite{cheng_large_2025}
proposed an  multimodal LLM(MLLM)-driven integrated sensing and communication (ISAC) framework and analyzed its beam-prediction performance.
However, current LLM-based methods have exhibited two primary limitations. First, they lack inherent \emph{spatial perception}, which is essential for 6G applications such as beamforming, blockage detection and dynamic resource allocation.\cite{yu_spatial-rag_2025} Spatial perception refers to the ability of a system to build and reason over a three-dimensional representation of its environment from multimodal sensor inputs. Without this capability, models may misinterpret critical geometric information, undermining vehicular network reliability. Second, these methods typically focus on a single task and fail to leverage the full potential of a unified backbone that can be utilized across multiple downstream objectives, sacrificing both efficiency and scalability.\\

\subsection{Contributions}  
The main contribution of this paper is a novel modular BEV‐injection connector that seamlessly integrates into any pre-trained LLM, enabling  3D multimodal spatial reasoning for V2I link performance prediction while eliminating the need for resource-intensive end-to-end retraining. This framework collects underexploited sensing data from neighboring vehicles and fuses them into a unified BEV representation. After that, we distill them with an instruction-guided Q-Former\cite{dai_instructblip_2023}, that dynamically selects the most task-relevant geometric features from the aggregated BEV map, reducing token overhead significantly compared to naive feature fusion approaches while preserving 3D spatial relationships critical for V2I scenarios. The resulting compact spatial tokens can be injected into any off-the-shelf LLM for accurate link quality assessment. Furthermore, this plug-and-play design also enables zero-shot generalization to unseen environmental conditions, maintaining performance advantage even under challenging nighttime and rainy scenarios. In summary, our key contributions include:

\begin{itemize}
  \item \textit{Plug-and-play modular BEV-injection connector:} We propose a lightweight, architecture-agnostic adapter that integrates multi-agent BEV features into the ego frame, and distill the instruction-relevant spatial cues in order to  inject them directly into the input of LLM for precise, context-driven proactive link assessment. 
  \item \textit{Cooperative BEV fusion for link-quality forecasting.} To the best of our knowledge, we are the first to incorporate a multi-agent collaborative scenario into a multimodal LLM framework for wireless communication tasks. 
  Our approach implements a temporal attention mechanism that fuses LiDAR point clouds and RGB images across distributed vehicular nodes and aligns them through precise coordinate-frame transformation.The framework's hierarchical BEV fusion pipeline effectively preserves geometric consistency across sensor inputs, enabling the frozen LLM to reason about wireless link quality with 3D spatial understanding. 
  \item \ti{V2I MLLM dataset:}   We develop a purpose-built dataset that combines high-fidelity CARLA simulations, MATLAB-based mmWave ray tracing, along  natural-language link-prediction queries with precise ground-truth labels extracted from ray-traced data to facilitate instruction-aware LLM training and evaluation in realistic multi-agent V2I communication scenarios.

\end{itemize}

Extensive experiments over our custom V2I multi-agent dataset show that the proposed BEV-injection connector improves the overall macro-average accuracy by 13.9 \% compared to an ego-only baseline. The results also show consistently improved performance across all tasks compared to the baseline. These results confirm that fusing multi-agent BEV maps fills ego blind spots, enriches geometric context, and enables the pre-trained LLM to  ``see around corners," filling critical blind spots in the ego vehicle's field of view. 

The rest of the paper is organized as follows. Section II details our system model. In Section III, we introduce our collaborative perception framework. Section IV presents simulation results and analysis. Finally, conclusions are drawn in Section V.

\section{System Model}
 
We consider a vehicle-to-infrastructure (V2I) scenario in a dynamic urban environment, where a set of vehicles $\mathcal{V}$ operate within the sensing and communication range of a roadside unit (RSU). Vehicle $v_0 \in \mathcal{V}$ is
designated as the \textit{ego vehicle} and communicates with the RSU over the wireless uplink. The remaining vehicles  serve as cooperative sensing agents and do not participate in communication. Each vehicle \( v \in \mathcal{V} \) is equipped with time-synchronized multimodal sensors, including multi-view RGB cameras, LiDAR, and GPS. 

At each discrete timestep \( t \in \mathcal{T} \), all vehicles transmit their sensor data to the RSU, , which fuses the multimodal inputs into a holistic three-dimensional representation of the environment. This representation is then processed by a multimodal large language model (MLLM) framework to predict key properties of the uplink between the RSU and the ego vehicle, such as signal quality, link stability, or anticipated degradation due to dynamic obstructions.

\subsection{Channel Model}

The uplink wireless channel between the ego vehicle \( v_0 \) and the RSU is modeled via a deterministic, geometry-based propagation framework. 
We employ ray tracing to capture complex multipath effects arising from the dense urban structure. The channel is characterized by a time-varying channel impulse response $h(t,\tau)$, which is expressed as:

\begin{equation}
h(t,\tau) = \sum_{l=1}^{L(t)} \alpha_l(t) e^{j\phi_l(t)} \delta(\tau - \tau_l(t)),
\end{equation}

\noindent
where $L(t)$ is the number of propagation paths at time $t$, $\alpha_l(t)$ is the amplitude, and $\phi_l(t)$ is the phase, and $\tau_l(t)$ is the propagation delay. For each ray path, the amplitude $\alpha_l(t)$ is calculated considering free-space path loss, reflection/transmission coefficients, and diffraction losses:

\begin{equation}
\alpha_l(t) = \frac{\lambda}{4\pi d_l(t)} \prod_{r \in \mathcal{R}_l} {\Gamma}_r \prod_{d \in \mathcal{D}_l} \mathcal{D}_d,
\end{equation}

\noindent
where $\lambda$ is the carrier wavelength, $d_l(t)$ is the total path length, $\Gamma_r$ is the reflection/transmission coefficient for each reflection point $r$ in the set of reflections $\mathcal{R}_l$ along the path, and $\mathcal{D}_d$ is the diffraction coefficient for each diffraction point $d$ in the set of diffractions $\mathcal{D}_l$. 
The ray-tracing framework captures V2I channel dynamics by summing contributions from LoS, reflected, and diffracted paths, and reflecting time-varying power levels and blockages caused by vehicles or urban structures.

\subsection{Multimodal Sensing Framework}

At each timestep \(t\), the RSU receives from each vehicle \(v_i\) the tuple
\[
\bigl(\mathcal{I}_i^t,\;\mathcal{L}_i^t,\;\mathcal{\xi}_i^t\bigr),
\]
where
\begin{itemize}
  \item \( \mathcal{I}_i^t = \{\bm{I}_{i,1}^t, \bm{I}_{i,2}^t, \ldots, \bm{I}_{i,N_c}^t\} \) are the \(N_c\) multi‐view RGB images,
  \item \(\mathcal{L}_i^t\)\;are the LiDAR point clouds,
  \item \(\mathcal{\xi}_i^t\)\;is the vehicle’s pose (position + orientation).
\end{itemize}

We fuse the RGB and LiDAR streams into a unified feature vector per agent.  Concretely, at time \(t\) for each vehicle \(v_i\) we have:
\begin{equation}
  \bm f_i
  = \phi_{\mathrm{enc}}\!\bigl(\bm I_i^t,\;\bm L_i^t\bigr)
  \;\in\;\mathbb R^d,
\end{equation}
where \(\phi_{\mathrm{enc}}\) is a frozen multimodal encoder that fuses the \(N_c\) camera views \(\bm I_i^t\) and the LiDAR point cloud \(\bm L_i^t\), and \(d\) is the shared embedding dimension.

These per-agent embeddings, together with their poses \(\xi_i^t\), are then fed into our trainable connector:
\begin{equation}
  \bm{z}^t
  = \phi_{\mathrm{conn, \bm {\theta}}}\!\bigl(\{\,(\bm f_i,\;\xi_i^t)\}_{i=1}^{N_v}\bigr)
  \;\in\;\mathbb R^{N_{\mathrm{tokens}}\times d_{\mathrm{LLM}}},
\end{equation}
where $\bm {\theta}$ represents the trainable parameters of the connector. The connector produces a compact token sequence \(\bm z_t\) that the frozen LLM can directly attend to (along with the language prompt).  Only \(\phi_{\mathrm{conn, \bm {\theta}}}\) is updated during training, while \(\phi_{\mathrm{enc}}\) and the LLM remain fixed.

Given a natural language instruction \( q^t \) (e.g., ``Is the communication link likely to be blocked in the next 3 time steps?"), the frozen LLM \( \mathcal{M}_{\text{LLM}} \) processes the fused token representation to generate a response:
\begin{equation}
    \hat{r}^t = \mathcal{M}_{\text{LLM}}(\bm{z}^t, q^t).
\end{equation}
Here, \( \hat{r}^t \) represents the predicted response (e.g., ``Yes, a large truck will likely block the line-of-sight path in approximately 2 time steps. "). The instruction \( q^t \) typically queries the LLM about the communication link status or channel conditions. 

The training objective is to update only the connector parameters \( \boldsymbol{\theta} \), while keeping all encoders and the LLM frozen. This is achieved by minimizing a task-specific loss over a dataset \( \mathcal{D} \):
\begin{equation}
    \min_{\boldsymbol{\theta}}
    \; \mathbb{E}_{(q^t, r^t) \sim \mathcal{D}}
    \Bigl[
      \mathcal{L}_{\mathrm{task}}
      \bigl(
        \mathcal{M}_{\mathrm{LLM}}(\bm{z}^t, q^t),
        r^t
      \bigr)
    \Bigr],
\end{equation}

\noindent
where \( r^t \) is the ground truth response.

Accurate V2I link prediction in dense urban environments must overcome rapid, unpredictable occlusions, fuse sensor data in real time, and generalize across lighting and weather without retraining from scratch.  Off-the-shelf multimodal LLMs, while powerful at language reasoning, lack explicit 3D spatial priors and cannot “see around corners.”  To bridge this gap, we introduce a plug-and-play BEV-injection connector that (i) preserves all frozen vision encoders and the LLM intact, (ii) uses underexploited sensing data from neighboring vehicles extending the field-of-view, and (iii) distills only the instruction-relevant spatial cues into a compact token set for the LLM to attend over.  This modular design leverages existing multi-agent data, reuses pretrained reasoning capabilities, and delivers substantial performance gains.
In the next section, we detail the architecture and of our 3D collaborative perception framework.

\addtolength{\topmargin}{0.02in}

\section{3D Collaborative Perception Framework}

\begin{figure}[t]
  \centering
  \includegraphics[width=\linewidth]{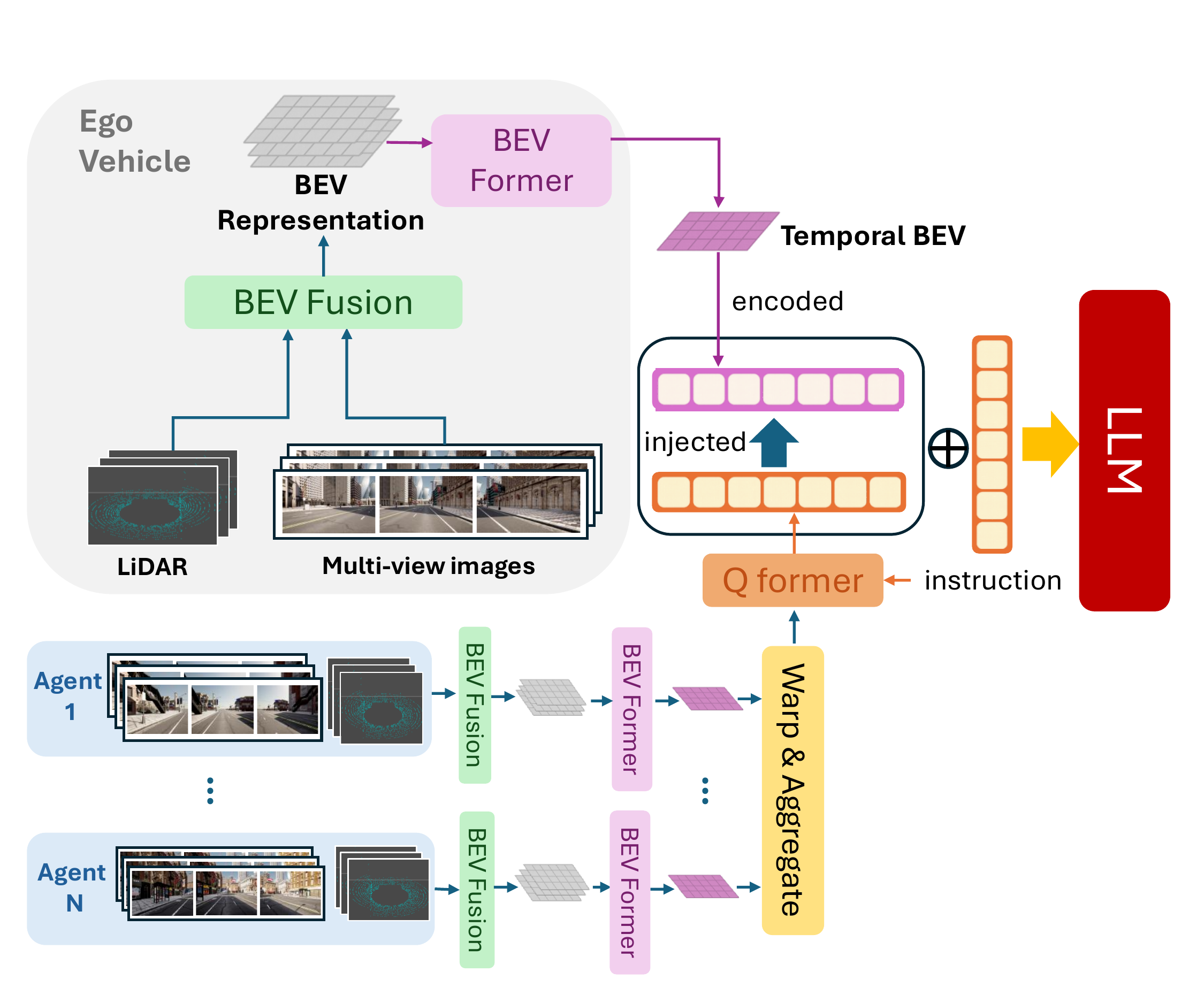}
  \caption{\small Overall architecture performs collaborative BEV fusion in the input stream of a frozen LLM to encourage spatial understanding.}
  \vspace{-3mm} 
  \label{fig:architecture}
\end{figure}

As shown in Figure \ref{fig:architecture}, our BEV-injection connector framework fuses ego-centric features with compact BEV tokens from cooperating vehicles to give a frozen MLLM genuine 3D awareness. By handling spatial alignment, temporal context, and multi-vehicle perspective in lightweight BEV modules, and then distilling only task-relevant cues via an instruction-aware Q-Former, we offload heavy reasoning from the LLM, yielding both efficiency and markedly improved link-quality prediction in cluttered environments.

At each time step \(t\), the ego vehicle’s \(N_c\) RGB cameras and LiDAR sweep $\bm{L}_t$ are fused into a unified BEV representation using the BEVFusion \cite{liu_bevfusion_2023} framework:
\begin{equation}
\bm{F}_{\mathrm{ego}}
= \operatorname{BEVFusion}\!\bigl(\{\bm{I}_t^{(j)}\}_{j=1}^{N_c},\,\bm{L}_t\bigr)\,.
\end{equation}
which captures both road topology and nearby obstacles from the ego’s viewpoint.  This fused BEV already improves on raw per‐view features by aligning modalities into a unified spatial grid.

To enable the model to reason about motion, we incorporate temporal fusion across a short sequence of consecutive frames over a fixed temporal window \( \{t - \Delta, \dots, t + \Delta\} \). To this end, we apply a BEVFormer\cite{li_bevformer_2022} temporal self attention (TSA) mechanism over the fused BEV sequence to produce a motion-aware BEV feature for vehicle \( v_i \):

\begin{equation}
\bm{B}^{\text{local}}_i = \text{TSA}\left(
\left\{
\text{BEVFusion}\left(
\bm{B}_{\text{img}}^{(\tau)},\,
\bm{B}_{\text{lidar}}^{(\tau)}
\right)
\right\}_{\tau = t - \Delta}^{t + \Delta}
\right)
\end{equation}

Since each BEV map is constructed in the local coordinate frame of its agent, we warp it into the ego vehicle's frame using relative GPS positions:
\begin{equation}
\tilde{\bm{B}}_i = \text{Warp}(\bm{B}^{\text{local}}_i,\, \xi_i \rightarrow \xi_{\text{ego}}).
\end{equation}
The warped BEVs are then fused using a $3 \times 3$ convolutional layer after channel-wise concatenation:
\begin{equation}
\bm{B}_{\text{agg}} = \text{Conv}_{3 \times 3} \left( \text{concat}(\tilde{\bm{B}}_1, \dots, \tilde{\bm{B}}_{N_v}) \right).
\end{equation}

\noindent
This multi‐agent aggregation fills in blind-spot regions and extends the field‐of‐view far beyond what a single LiDAR scan can see. The raw BEV tensors are too large to be used as direct LLM input. Therefore, we distill only the relevant spatial cues by applying an instruction-aware Q-Former \cite{dai_instructblip_2023} to the aggregated BEV map:
\begin{equation}
\bm{F}_{\text{bev}} = \phi_{\text{QF}}([\bm{Q}_{\text{bev}};\, \bm{L}_{\text{inst}}],\, \bm{B}_{\text{agg}}).
\end{equation}
These BEV tokens are fused with the visual stream through cross-attention:
\begin{equation}
\bm{F}'_{\text{ego}} = \bm{F}_{\text{ego}} + \text{CrossAttn}(\bm{F}_{\text{ego}}, \bm{F}_{\text{bev}}).
\end{equation}

\noindent
Finally, the model composes its response using the LLM:
\begin{equation}
\hat{r}_t = \text{LLM}([\bm{L}_{\text{inst}};\, \bm{F}'_{\text{ego}};\, \bm{F}_{\text{bev}}]).
\end{equation}

\noindent
 In practice, this design: 1) dramatically improves blockage handling by collaborative spatial perception via neighbor BEVs. 2) Focuses the LLM’s attention on task‐relevant spatial cues, avoiding information overload. 3) Preserves pretrained language and vision knowledge by keeping large backbones frozen.






\begin{algorithm}[t]
\caption{Collaborative BEV‐injected LLM Inference Framework}\label{alg:compact}
\DontPrintSemicolon
\SetNlSkip{0.9em}              
\setlength{\algomargin}{0.8em} 

\begin{minipage}{0.94\linewidth}  

\KwIn{Ego images $\{\bm I_{0,j}\}$, LiDAR $\bm L_0$,\\
      Helper data $\{\bm I_i,\bm L_i,\xi_i\}_{i=1}^{N_v}$,\\
      Instruction tokens $\bm L_{\mathrm{inst}}$}
\KwOut{Predicted response $\hat r$}

\smallskip
\textbf{1. Ego BEV Fusion:}\;
$\bm F_{\mathrm{ego}}\leftarrow 
  \mathrm{BEVFusion}(\{\bm I_{0,j}\},\,\bm L_0)$\;

\smallskip
\textbf{2. Helper BEV Aggregation:}\;
\For{$i\leftarrow1$ \KwTo $N_v$}{
  $\bm B^{\mathrm{loc}}_i\leftarrow 
     \mathrm{TSA}\!\bigl(\mathrm{BEVFusion}(\bm I_i,\,\bm L_i)\bigr)$\;
  $\tilde{\bm B}_i\leftarrow 
     \mathrm{Warp}(\bm B^{\mathrm{loc}}_i,\,\xi_i\!\to\!\xi_{\mathrm{ego}})$\;
}
$\bm B_{\mathrm{agg}}\leftarrow
  \mathrm{Conv}_{3\times3}\bigl[\tilde{\bm B}_1,\dots,\tilde{\bm B}_{N_v}\bigr]$\;

\smallskip
\textbf{3. Instruction‐Aware Distillation:}\;
$\bm F_{\mathrm{bev}}\leftarrow 
  \phi_{\mathrm{QF}}\bigl([\bm Q_{\mathrm{bev}};\,\bm L_{\mathrm{inst}}],\,
                       \bm B_{\mathrm{agg}}\bigr)$\;

\smallskip
\textbf{4. BEV Injection \& Reasoning:}\;
$\bm F'_{\mathrm{ego}}\leftarrow 
  \bm F_{\mathrm{ego}}
  + \mathrm{CrossAttn}(\bm F_{\mathrm{ego}},\,\bm F_{\mathrm{bev}})$\;
$\hat r \leftarrow 
  \mathrm{LLM}\bigl([\bm L_{\mathrm{inst}};\,
                  \bm F'_{\mathrm{ego}};\,\bm F_{\mathrm{bev}}]\bigr)$\;

\end{minipage}
\end{algorithm}

\section{Simulation Results and Analysis}

\begin{figure}[!t]
  \centering
  \includegraphics[width=\linewidth]{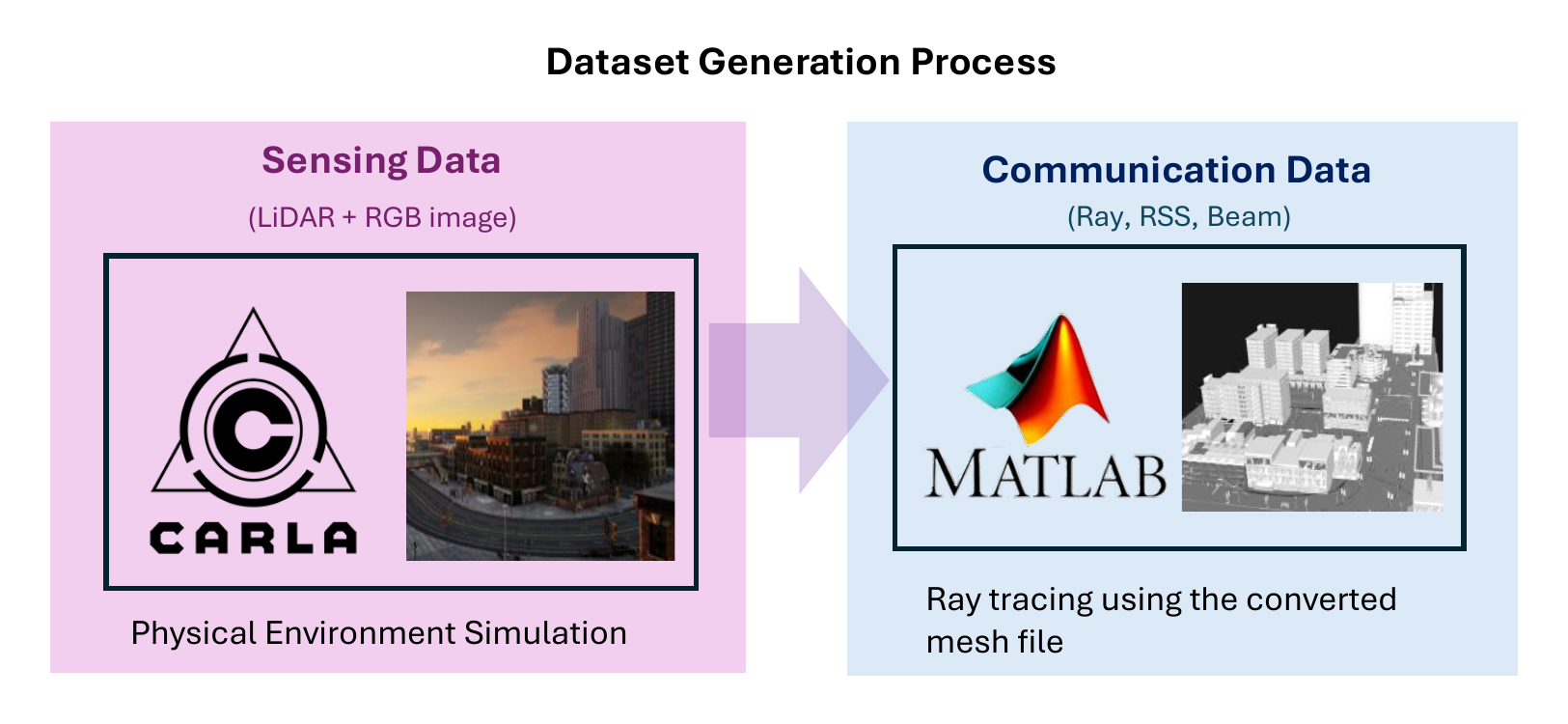}
  \caption {\small Data generation using CARLA with MATLAB mmWave ray tracing.}

  \vspace{-3mm} 
  \label{fig:architecture}
\end{figure}

\subsection{Data Generation}

To train the BEV-fusion connector, we developed a co-simulation framework that integrates the autonomous driving
simulator CARLA \cite{dosovitskiy_carla_2017} with MATLAB-based mmWave ray
tracing, inspired by \cite{park_resource-efficient_2025}. The dataset
includes 50 episodes in the Town 10 map, each lasting up to
200 frames sampled every 100 milliseconds from five cooperative vehicles and a single RSU. At each frame, every agent
captures three synchronized RGB views, a LiDAR sweep, and
GPS poses. The ray tracing framework outputs per-frame received power values and ray data. To increase diversity, 30 episodes occur at noon, 10 at night, and 10 in rain.
We automatically generate natural-language link prediction
queries and groundtruth responses  from the ray-traced data.
\subsection{Setup}

We use the Llama-3.2-11B-Vision \cite{noauthor_llama_nodate}  model as our LLM backbone.  The BEV-fusion connector are trained while keeping both the LLM and its vision encoder frozen.

Training is performed on our custom V2I dataset. Our dataset was partitioned into training (80\%), validation (10\%), and test (10\%) subsets. To generate frame sequences,  three key frames were sampled in sequence for each episode. We used 5 sensing agent vehicles in each scenario. The AdamW~\cite{loshchilov2019decoupledweightdecayregularization} optimizer was used for training with a weight decay of 0.05. A cosine scheduler is used, starting at $10^{-4}$ with linear warm-up over the first 5\% of steps. We use a batch size of 16 and train the connector for 15 epochs.

\subsection{Task Definitions}

We evaluate our framework across three cooperative link‐prediction tasks, defined precisely as follows:

\begin{enumerate}
  \item \ti{LoS/NLoS classification:}  
    Determine whether the path between the RSU and the ego vehicle is clear (LoS) or blocked (NLoS).

  \item \ti{Link availability classification:}  
    Classify the link as available if the received signal strength is at least \(-80\)\,dBm, otherwise unavailable. This threshold ensures sufficient SNR to maintain QPSK modulation over 100 MHz bandwidth channels.

  \item \ti{Blockage risk prediction:}  
    Predict whether the link will transition from clear to blocked within the next $3$ time steps.
\end{enumerate}
\subsection{Quantitative Results and Analysis}
Table~\ref{tab:qual_results} evaluates the performance on our three cooperative classification tasks, where we additionally compute a macro-average of accuracy across these tasks to summarize overall gains. We compare BEV-injection against non-LLM baselines retrained per task: (i) a 3-layer LSTM \cite{hochreiter_long_1997}, (ii) a 3-layer GRU \cite{chung_empirical_2014}, and (iii) a 4-layer Transformer encoder \cite{vaswani_attention_2023}. Each non-LLM baseline processes the same multimodal ego-vehicle inputs (RGB + LiDAR) through feature extractors, followed by task-specific classification heads that are retrained from scratch for individual tasks. In contrast, our approach leverages a single frozen LLM backbone that generalizes across all three tasks without requiring task-specific retraining, demonstrating the versatility of instruction-guided multimodal reasoning. Despite this unified architecture, our method outperforms all task-optimized non-LLM baselines, confirming that collaborative BEV injection combined with an LLM enables both more effective and adaptive spatial reasoning compared conventional approaches.
The results in Table~\ref{tab:qual_results} also confirm that explicitly injecting BEV representations into the input of a frozen LLM backbone yields a qualitatively different reasoning capability compared to ego-only models.  Across all three classification tasks, performance gains are largest for those requiring precise spatial understanding, specifically distinguishing line-of-sight versus non-line-of-sight and predicting blockages.  These tasks demand not only local appearance cues but also geometric context spanning multiple viewpoints. The BEV tokens distilled by our Q-Former supply this, allowing the model to “see around corners” by aggregating complementary LiDAR and image information from helpers.

\newcommand{\NA}{\multicolumn{1}{c}{—}}
\begin{table*}[t]
  \centering
  \small
  \setlength{\tabcolsep}{4.5pt}
  \sisetup{table-format=2.1,detect-weight,detect-family}
  \caption{Performance on cooperative link prediction tasks. We compare three ego-only baselines and three non-LLM baselines against our BEV-injection model across line-of-sight detection, link availability classification, and blockage risk prediction. The macro-F1 score is simply the average of the F1 scores computed separately for each class.}
  \label{tab:qual_results}
  \resizebox{\textwidth}{!}{%
  \begin{tabular}{@{} ll
                  *{3}{S}  
                  *{3}{S}  
                  S        
                  @{}}
    \toprule
    \textbf{Task} & \textbf{Metric}
      & \multicolumn{3}{c}{\textbf{Ego-only (LLM baseline)}} 
      & \multicolumn{3}{c}{\textbf{Ego-only (Non-LLM heads)}}
      & \textbf{BEV Injection (Proposed)} \\
    \cmidrule(lr){3-5}\cmidrule(lr){6-8}
      & &
      \textbf{Img+LiDAR} & \textbf{LiDAR} & \textbf{Image}
      & \textbf{LSTM} & \textbf{GRU} & \textbf{Transformer}
      & \\
    \midrule
    Line-of-sight vs.\ non-line-of-sight
      & Accuracy   & 67.2 & 61.0 & 54.5 & 72.1 & 71.8 & 74.2 & \textbf{83.1} \\
      & Macro-F1   & 65.5 & 59.2 & 52.3 & 70.3 & 70.0 & 72.1 & \textbf{81.3} \\
    \addlinespace
    Link availability
      & Accuracy   & 78.3 & 72.1 & 65.8 & 79.5 & 79.1 & 81.2 & \textbf{90.1} \\
      & F1 Score   & 76.2 & 69.5 & 62.7 & 77.8 & 77.4 & 79.5 & \textbf{88.9} \\
    \addlinespace
    Blockage risk prediction
      & Accuracy   & 74.5 & 68.8 & 62.0 & 76.2 & 75.8 & 77.5 & \textbf{88.5} \\
      & Precision  & 75.0 & 73.2 & 66.4 & 77.1 & 76.7 & 78.9 & \textbf{92.4} \\
      & Recall     & 72.5 & 71.3 & 64.8 & 74.8 & 74.3 & 76.2 & \textbf{89.2} \\
    \midrule
    \textbf{Overall Macro-Avg.}
      & Accuracy   & 73.3 & 67.3 & 60.8 & 76.0 & 75.6 & 77.6 & \textbf{87.2} \\
    \bottomrule
  \end{tabular}%
  }
\end{table*}

\begin{table}[!t]
  \centering
  \small
  \sisetup{
    table-format=2.1,
    table-space-text-post={\%},
    detect-weight,
    detect-family
  }
  \caption{Ablation study on BEV-injection components. Each row removes a single module to quantify its impact on the three tasks and overall macro-average accuracy.}
  \label{tab:ablation}
  \resizebox{\columnwidth}{!}{%
    \begin{tabular}{@{} l S S S S @{}}
      \toprule
      \textbf{Variant}
        & {\textbf{LoS/NLoS} (Acc)}
        & {\textbf{Link Avail} (F1)}
        & {\textbf{Blockage} (Acc)}
        & {\textbf{Macro-Avg} (Acc)} \\
      \midrule
      Ego-only baseline
        & 67.2 & 76.2 & 74.5 & 73.3 \\
      w/o Temporal Fusion
        & 78.0 & 86.0 & 83.0 & 82.3 \\
      w/o Q-Former
        & 80.0 & 87.0 & 85.0 & 84.0 \\
      w/o Multi-agent Warp
        & 76.0 & 84.0 & 82.0 & 80.7 \\
      \midrule
      Full BEV-injection
        & 83.1 & 88.9 & 88.5 & 87.2 \\
      \bottomrule
    \end{tabular}
  }
\end{table}

\begin{figure}[t]
  \centering
  \includegraphics[width=\linewidth]{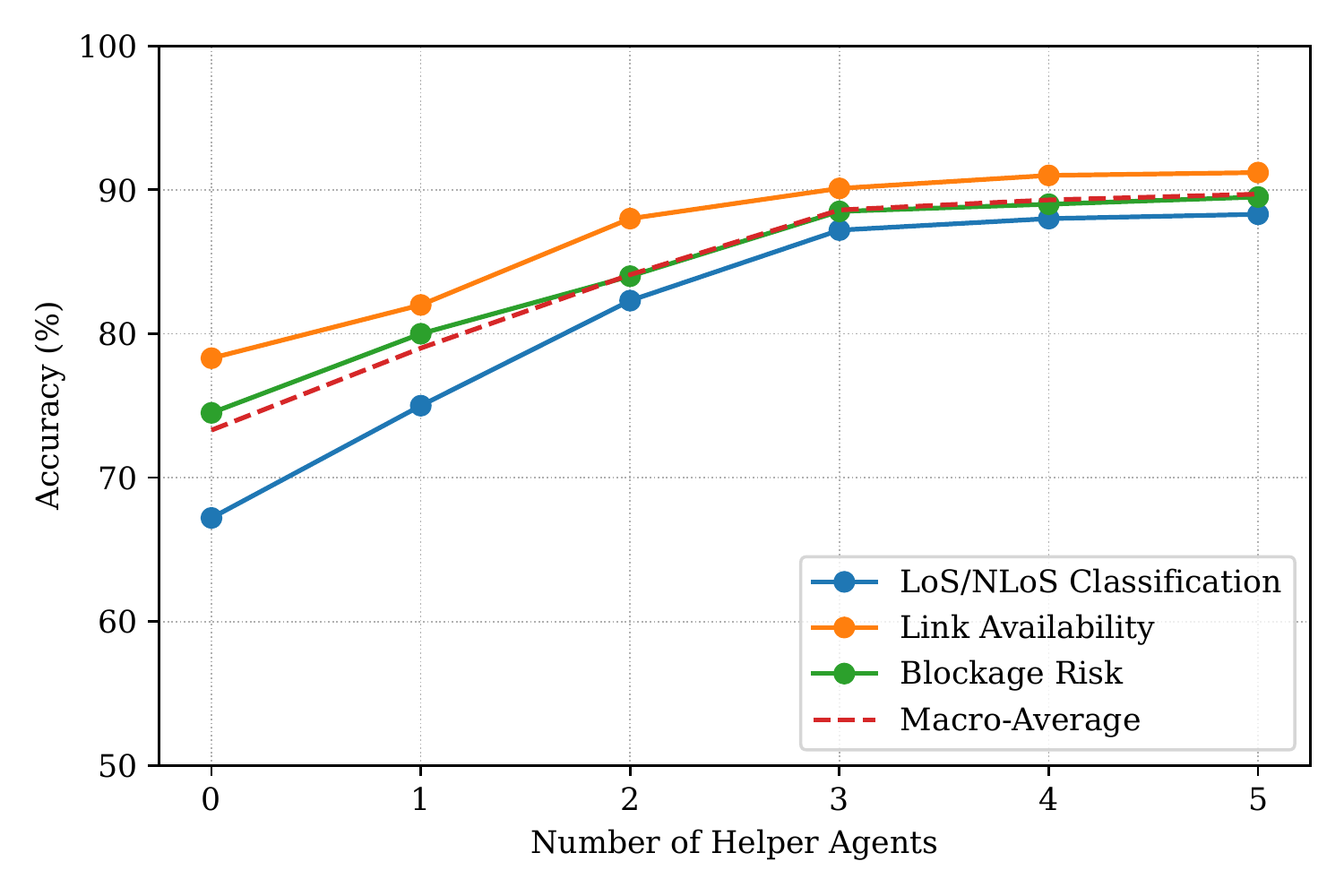}
  \caption{\small Effect of increasing the number of helper vehicles on macro-average accuracy. The initial helpers yield the largest gains, while additional agents provide diminishing returns.}
  \vspace{-3mm}
  \label{fig:acc_vs_agents}
\end{figure}

\begin{figure}[t]
  \centering
  \includegraphics[width=\linewidth]{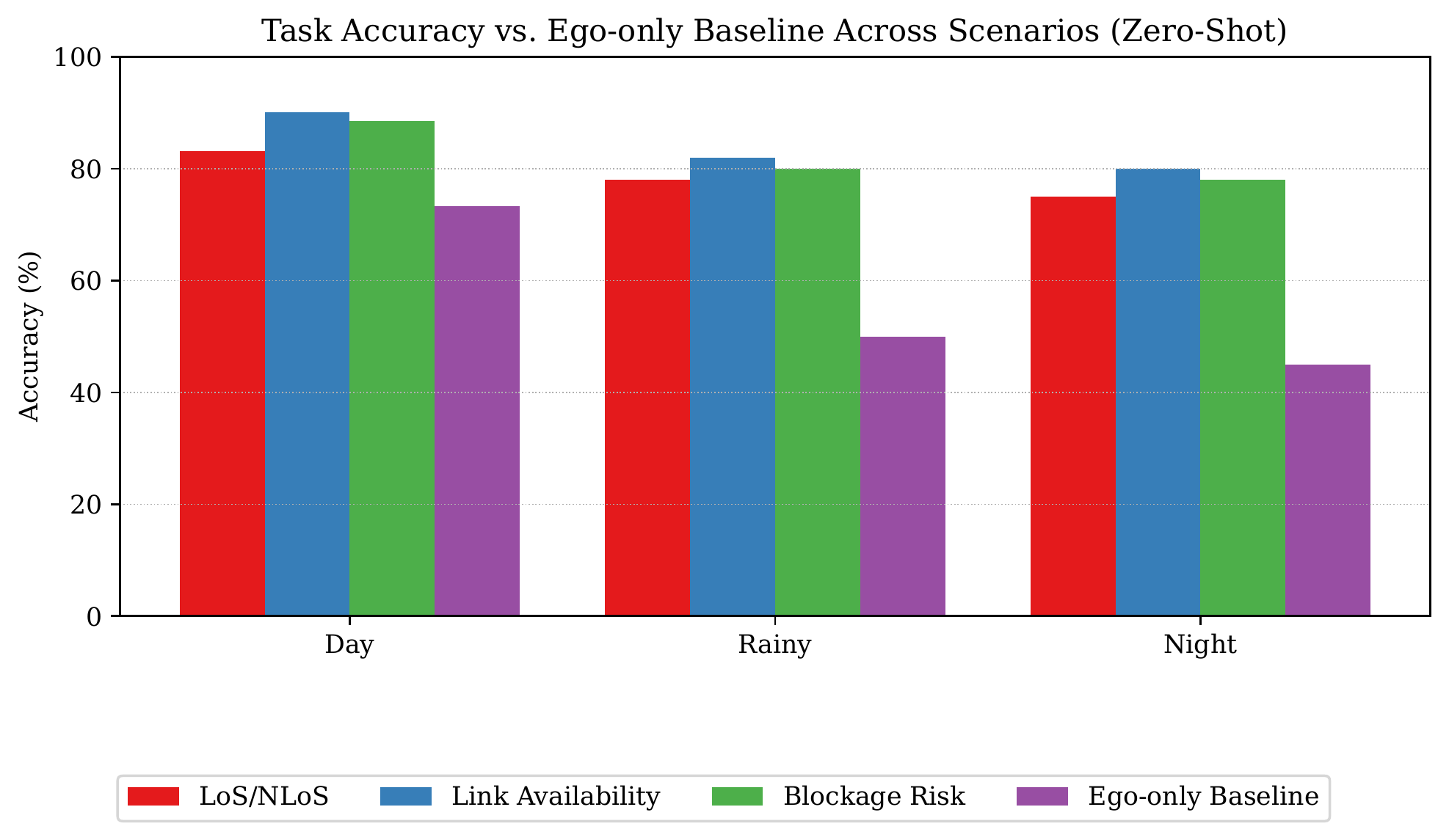}
  \caption{ \small Zero-shot generalization from Day training to Rain and Night scenario testing. The BEV-injection model sustains a significant performance margin under domain shift compared to ego-only baselines.}
  \vspace{-5mm} 
  \label{fig:generalization}
\end{figure}

Figure~\ref{fig:acc_vs_agents} shows the accuracy of our V2I link‐quality prediction tasks as a function of the number of helper agents, where a helper agent is a neighboring vehicle that shares its local sensor data for cooperative BEV construction. In this figure, we observe that one or two neighbors rapidly improve accuracy by covering blind spots, while additional agents give smaller gains. This indicates that even a couple of well-positioned vehicles can fill critical blind spots in the ego’s field of view.  

Figure~\ref{fig:generalization} shows zero-shot generalization from the clear daytime training set to unseen rainy and nighttime conditions.  While achieving impressive daytime performance, the BEV-injection model truly distinguishes itself under adverse conditions. In rainy scenarios, our method maintains robust 80.0\% macro-average accuracy, whereas the ego-only model catastrophically deteriorates by 23 points to a barely useful 50.0\%. Similarly, in nighttime scenarios our approach sustains a  77.7\% macro-average while the baseline collapses to an unacceptable 45.0\%. This exceptional domain robustness stems from our architecture's fundamental advantage: by reasoning over geometry-focused BEV tokens rather than raw pixel intensities, the LLM can interpret spatial relationships consistently across environmental variations.

 Table~\ref{tab:ablation} presents an ablation study to shed  light on each component’s contribution.  Removing temporal fusion degrades the model’s ability to capture short-term motion cues critical for blockage forecasting, while skipping the Q-Former harms the distillation of relevant spatial features into BEV queries.  Finally, omitting the multi-agent warp step misaligns helpers’ BEV maps, erasing the benefits of coordinate consistency and leading to a marked drop in all task metrics.

\section{Conclusion}
In this paper, we have developed a novel BEV-injection framework that endows MLLMs with the three-dimensional spatial reasoning required for reliable V2I link performance prediction. By aggregating passive multi-view RGB and LiDAR data from neighboring vehicles into a shared BEV representation, distilling it through an instruction-aware Q-Former, and injecting the resulting spatial tokens into the frozen MLLM, the proposed approach bridges the gap between language-driven reasoning and precise spatial context. Coupled with a purpose-built V2I dataset, this method significantly outperforms ego-only baselines by 13.9\% in daytime scenarios and by up to 32.7\% in rainy and nighttime conditions, demonstrating robustness to environmental challenges.

\bibliographystyle{IEEEtran}
\bibliography{refmain,ref}

\end{document}